\documentclass[letterpaper, 10 pt, conference]{IEEEtran}  % Comment this line out if you need a4paper
\usepackage[T1]{fontenc}
\usepackage{algorithm}
\usepackage{algpseudocode} % or algorithmic

\usepackage{xcolor}
\usepackage{upgreek}
\usepackage{multicol}
\usepackage{dsfont,setspace}
\usepackage{amsmath,graphicx}
\usepackage{float}
\usepackage{amssymb}
\usepackage{amsmath}
\usepackage[normalem]{ulem}
\usepackage[thinc]{esdiff}

\usepackage{lineno}
\usepackage{epstopdf}
\usepackage{authblk}
\usepackage{dblfloatfix}
\usepackage{tabularx}
\usepackage{soul}

\usepackage[
  separate-uncertainty = true,
  multi-part-units = repeat
]{siunitx}

\usepackage{subcaption}
\usepackage[colorlinks]{hyperref}
\hypersetup{
     colorlinks,
     linkcolor=blue,
     citecolor=blue,
     filecolor=black,
     urlcolor=blue,
 } 
\usepackage{cleveref}
\usepackage{textcomp}
\usepackage{gensymb}
\usepackage{tikz}
\usepackage{pifont}% http://ctan.org/pkg/pifont
\usepackage{makecell}
\usepackage{svg}
\usepackage{multirow}

\newcommand{\grey}[1]{
{\color{gray}{#1}}
}

\IEEEoverridecommandlockouts                              % This command is only needed if 
\begin{document}

\title{\LARGE \bf VIP-Loco: A Visually Guided Infinite Horizon Planning Framework for Legged Locomotion}

\author{
% Anonymous Authors
Aditya Shirwatkar$^{1}$, Satyam Gupta $^{1}$, Shishir Kolathaya$^{2}$
\thanks{This work is funded by Kotak IISc AI-ML Centre (KIAC) - Google Grant}
\thanks{$^{1}$A. Shirwatkar, and S. Gupta are with the Robert Bosch Center for Cyber-Physical Systems, Indian Institute of Science, Bengaluru.}%
\thanks{$^{2}$S. Kolathaya is with the Robert Bosch Center for Cyber-Physical Systems and the Department of Computer Science \& Automation, Indian Institute of Science, Bengaluru.}
\thanks{
Project Website: \href{https://www.stochlab.com/VIP-Loco/}{stochlab.com/VIP-Loco/}, 
Email: \href{mailto:stochlab@iisc.ac.in}{stochlab@iisc.ac.in}}%
% \thanks{
% }
}

\maketitle
\thispagestyle{empty}
\pagestyle{empty}

%%%%%%%%%%%%%%%%%%%%%%%%%%%%%%%%%%%%%%%%%%%%%%%%%%%%%%%%%%%%%%%%%%%%%%%%%%%%%%%%
\begin{abstract}
Perceptive locomotion for legged robots requires anticipating and adapting to complex, dynamic environments. 
Model Predictive Control (MPC) serves as a strong baseline, providing interpretable motion planning with constraint enforcement, but struggles with high-dimensional perceptual inputs and rapidly changing terrain. 
In contrast, model-free Reinforcement Learning (RL) adapts well across visually challenging scenarios but lacks planning. 
To bridge this gap, we propose VIP-Loco, a framework that integrates vision-based scene understanding with RL and planning. 
During training, an internal model maps proprioceptive states and depth images into compact kinodynamic features used by the RL policy.
At deployment, the learned models are used within an infinite-horizon MPC formulation, combining adaptability with structured planning. 
We validate VIP-Loco in simulation on challenging locomotion tasks, including slopes, stairs, crawling, tilting, gap jumping, and climbing, across three robot morphologies: a quadruped (Unitree Go1), a biped (Cassie), and a wheeled-biped (TronA1-W). 
Through ablations and comparisons with state-of-the-art methods, we show that VIP-Loco unifies planning and perception, enabling robust, interpretable locomotion in diverse environments.
\end{abstract}


%%%%%%%%%%%%%%%%%%%%%%%%%%%%%%%%%%%%%%%%%%%%%%%%%%%%%%%%%%%%%%%%%%%%%%%%%%%%%%%%
\section{Introduction}

Model Predictive Control (MPC) \cite{cvx_mpc, jumpmpc} and Reinforcement Learning (RL) \cite{rma, dreamwaq, himloco} have driven major progress in legged locomotion, enabling robust performance across diverse terrains. MPC offers interpretable control by optimizing actions over a horizon under constraints, while RL achieves adaptability and generalization through policy optimization. 
Despite these strengths, a majority of the methods in both paradigms exhibit reactive behavior when relying on proprioceptive observations, which limits their performance in scenarios requiring foresight, such as gap crossing, confined navigation, or high-penalty obstacle traversal.

\begin{figure}[htp!]
    \scriptsize
    \captionsetup{font=footnotesize}
    \centering
    \includegraphics[width=0.6\linewidth]{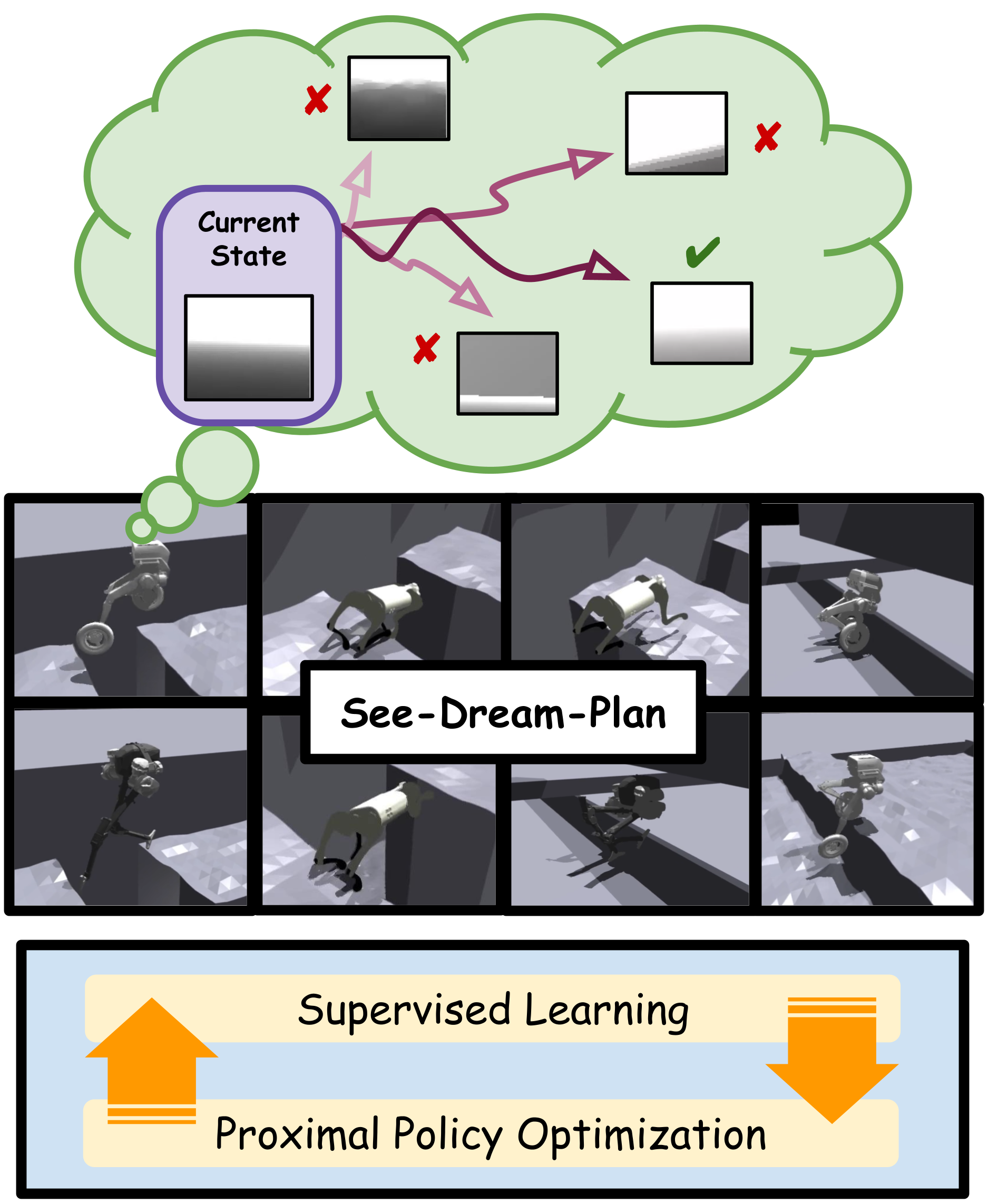}
    % \caption{\textit{Conceptual outline of VIP-Loco} -- The framework integrates vision and proprioception to guide locomotion. During training, an internal model predicts compact predictive representations from depth images and proprioceptive inputs, enabling the RL policy and critic. At deployment, an approximate infinite-horizon MPC module iteratively refines action trajectories, ensuring long-term reward maximization while satisfying kinodynamic and safety constraints.}
    \caption{{\textit{Conceptual outline of VIP-Loco} -- The framework learns a compact internal model from vision and proprioception during training, which is then utilized by an infinite-horizon MPC planner at deployment to enable anticipatory, constraint-aware locomotion.}}
    \label{fig:overview}
    \vspace{-15pt}
\end{figure}

\begin{figure*}
    \vspace*{0.2cm}
    \scriptsize
    \captionsetup{font=footnotesize}
    \centering    
    \includegraphics[width=0.8\linewidth]{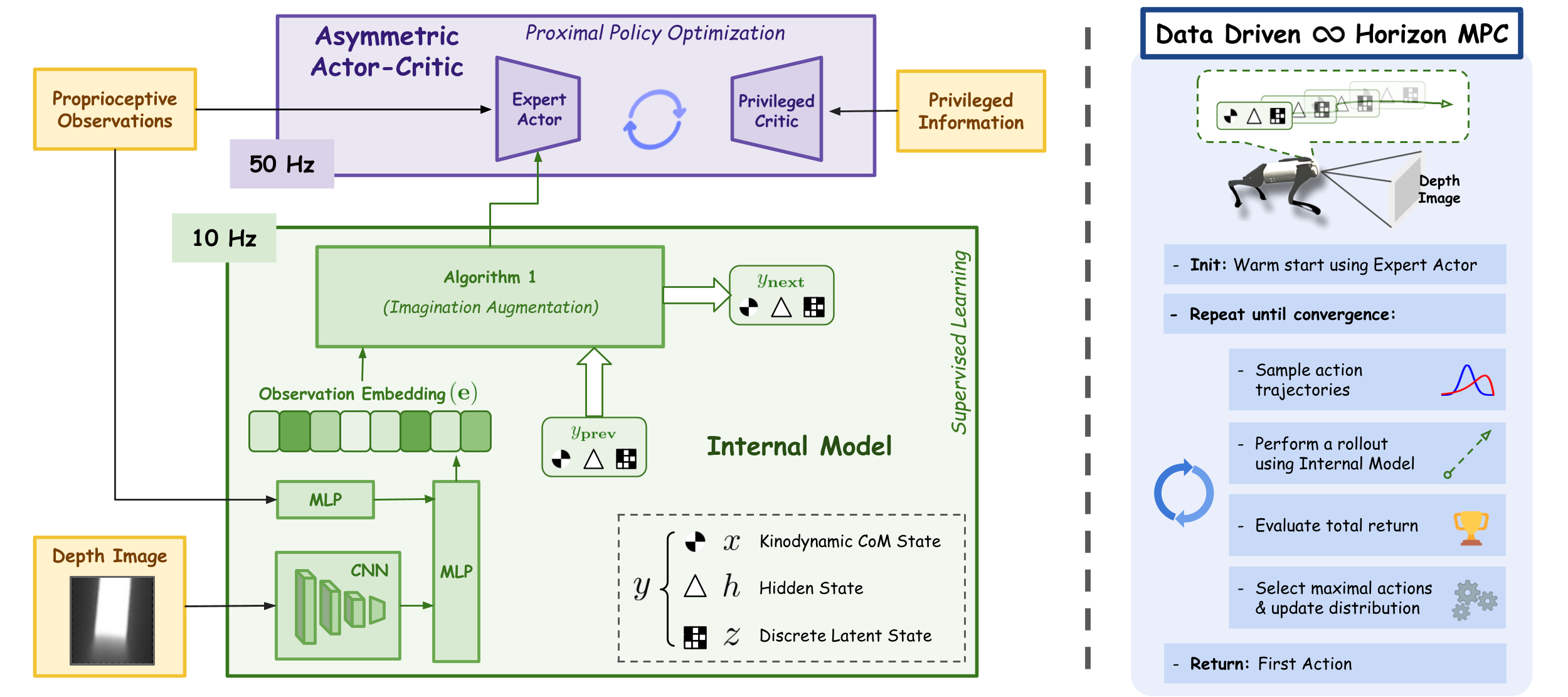}
    \caption{
    % \textit{Overview of VIP-Loco Framework --} The proposed framework consists of two major components: (1) \textit{Learning Stage (left)} – an asymmetric actor-critic architecture trained with PPO using proprioceptive observations, internal model augmentations, and privileged information. An internal model, trained via supervised learning, predicts kinodynamic states, recurrent and latent dynamics, using both depth images and proprioceptive features. Imagination augmentation enables the actor to reason about future states using future rollouts. 
    {\textit{Overview of VIP-Loco Framework --} The proposed framework consists of two major components: (1) \textit{Learning Stage (left)} – The internal model includes a GRU cell $g_\varphi$ (operating at 10~Hz, updating recurrent memory $h$ at each step) coupled with a CNN-MLP encoder for depth processing, an encoder/dynamics pair for latent state estimation, and reward/value heads. The Expert Actor (50~Hz) receives the imagined rollout $\mathcal{X}$ and hidden state $h$ via stop-gradient.}
    (2) \textit{Deployment Stage (right)} – a data-driven MPC that uses the learned internal model to iteratively sample and refine trajectories, leveraging vision-based scene understanding to select actions that maximize long-term reward while satisfying kinodynamic constraints.}
    \label{fig:training}    
    \vspace{-15pt}
\end{figure*}

Legged robots operating in real-world environments must anticipate upcoming terrain variations and plan motion trajectories that satisfy stability and kinodynamic constraints. Purely reactive controllers often fail in such settings because they lack predictive reasoning. This gap has motivated research into integrating perception and planning for locomotion. 
{Recent works have begun incorporating exteroceptive sensing to extend beyond purely reactive control ~\cite{pie, wmp, piploco, himloco, parkourcmu, parkoureth, parkourstanford, egocentric, multi_brain}. We discuss these lines of work in detail in Section~\ref{sec:related}, yet a unified framework combining visual foresight with planning has remained elusive. 
}

Thus, a key challenge lies in unifying perception, long-horizon predictive control, and real-time deployment. An ideal framework must leverage rich visual context for terrain understanding, maintain interpretability through structured planning, and remain computationally feasible for onboard control.
To bridge these gaps, we introduce VIP-Loco, a Visually Guided Infinite-Horizon Planning framework for Legged Locomotion {(see Fig. \ref{fig:overview} for a conceptual overview)}.
Our approach integrates vision-based scene understanding with an internal model-driven predictive control, enabling robots to anticipate terrain variations and reason over extended horizons while maintaining real-time feasibility. 
{We show that this achieves robust, adaptive locomotion across challenging environments that require proactive decision-making.}

In summary, our contributions are:
\begin{itemize}
    \item \textbf{Visually-Guided Infinite-Horizon Planning:} {We use a visually guided internal model-driven predictive control for legged locomotion, where infinite-horizon return is approximated via value function bootstrapping at the terminal planning state.}
    
    \item \textbf{Interpretable Internal Model with Kinodynamic Predictions:} Our internal model predicts kinodynamic states in compact representations for imagination augmentation, bridging the gap between perception-driven reinforcement learning and structured planning.
        
    \item \textbf{Extensive Multi-Morphology Evaluation:} We demonstrate our approach across three distinct morphologies: a quadruped (Unitree Go1), a biped (Cassie), and a wheeled-biped (TronA1-W), and validate it on challenging terrains such as stairs, gaps, crawling, narrow passages, and climbing.
\end{itemize}

% \red{We now present the details of our proposed method, VIP-Loco, starting with the related work, the methodology, and the results.
% }

{
\section{Related Work}
\label{sec:related}

\textbf{Proprioceptive Locomotion with RL.}
Reinforcement learning has enabled robust locomotion across diverse terrains using proprioceptive inputs alone ~\cite{dreamwaq, himloco}. Teacher-student distillation ~\cite{rma, eth2020} can transfer privileged-information policies to deployable ones, achieving surprisingly capable blind locomotion. However, these methods fundamentally lack visual foresight, limiting performance in scenarios requiring anticipation, such as gap crossing or confined navigation.

\textbf{Perceptive Locomotion with RL.}
Incorporating depth, LiDAR, or RGB inputs has enabled more agile, terrain-aware behaviors ~\cite{egocentric,parkoureth,parkourstanford,pie,multi_brain,dreamwaq++}. ANYmal Parkour~\cite{parkoureth} and Robot Parkour Learning~\cite{parkourstanford} demonstrate that perception-conditioned RL policies can learn highly dynamic maneuvers. WMP~\cite{wmp} introduces a world model to learn recurrent visual representations for locomotion, achieving strong generalization. However, these methods do not employ explicit planning at deployment, relying only on the learned policy to implicitly 
encode foresight.

\textbf{MPC and Hybrid Frameworks.}
Classical MPC~\cite{cvx_mpc,nmpc} provides interpretable, constraint-satisfying control but struggles with high-dimensional visual inputs and generalization. Infinite-horizon MPC variants~\cite{loop, mopac, dmdmpc, d3p, rilqr, tdmpc, tdmpc2} combine learned dynamics and value models with trajectory optimization, but typically rely on off-policy training pipelines ill-suited to massively parallel simulation. PIP-Loco~\cite{piploco} bridges on-policy RL training with MPC deployment for proprioceptive locomotion. 

For VIP-Loco, we aim to extend these paradigms of perception-based locomotion (WMP~\cite{wmp}) and MPC (PIP-Loco~\cite{piploco}) to enable anticipatory planning in visually complex environments, while maintaining on-policy training efficiency and addressing environments not addressed by either predecessor alone.
% \textbf{Positioning.}
% The closest prior work to ours is PIP-Loco~\cite{piploco} (proprioceptive, 
% with planning) and WMP~\cite{wmp} (visual, without planning). VIP-Loco occupies the intersection of visual inputs and explicit infinite-horizon planning. 
% The ablation structure in Section~\ref{sc:results} is designed to quantify the contribution of each axis directly.
}

\section{methodology} \label{sc:method}
This section will describe our proposed method, starting with some preliminary concepts. 

\subsection{Setup}
We formulate legged locomotion as a Partially Observable Markov Decision Process (POMDP), represented by the tuple $\mathcal{M} = (\mathcal{S}, \mathcal{A}, \mathcal{O}, r, \mathcal{P}, \gamma, \mathcal{F})$. The system state space is denoted as $\mathcal{S} \subset \mathbb{R}^n$, while the agent receives observations from an observation space $\mathcal{O} \subset \mathbb{R}^p$ ($p << n$). The control inputs belong to an action space $\mathcal{A} \subset \mathbb{R}^m$. Additionally, we define an observation history $\mathcal{O}^M \subset \mathbb{R}^{p \times M}$, which aggregates sensory information over the past $M$ steps. Observations are obtained via the mapping $\mathcal{F}: \mathcal{S} \mapsto \mathcal{O}$. 
The dynamics of the environment are governed by a transition function $\mathcal{P}: \mathcal{S} \times \mathcal{A} \mapsto \Pr(\mathcal{S})$, where $\Pr$ represents a probability distribution over future states. The agent's objective is to maximize the cumulative discounted reward, where the reward function is given by $r: \mathcal{S} \times \mathcal{A} \to \mathbb{R}$, and future rewards are discounted by a factor $\gamma \in (0,1)$. 

In this work, to effectively solve this POMDP, we employ an Asymmetric Actor-Critic \cite{asymmetric} architecture. While the expert actor and privileged critic are trained via the Proximal Policy Optimization (PPO) \cite{ppo}, the internal model is trained via supervised learning (see Fig. \ref{fig:training}). The exact details of this will be discussed in the later subsections.

\subsubsection{Observation Space} At each time step $t$, the observation $\mathbf{o}_t$ comprises a $5$-step history of the robot's joint angles relative to their nominal values, joint velocities, projected gravity, target velocity commands for the robot's base, and the previous action. We collectively refer to these as proprioceptive observations. To incorporate vision, we introduce additional inputs of $2$D depth images in $\mathbf{o}_t$. Finally, we denote the set of privileged observations $\mathbf{o}^\mathrm{priv}_t$, which consists of proprioceptive terms and some extra terms like scan dots that provide height field information around the robot, base linear velocities, reaction forces, mass, and friction parameters.

\subsubsection{Action Space} In the case of quadruped and biped, the policy outputs joint-angle perturbations relative to the nominal joint angles for all $12$ degrees of freedom (DoF). For the wheeled-biped, the action also includes joint-velocity commands for the wheel DoF. The resulting joint command is given to a PD controller, which computes the corresponding torques to achieve the desired motion.

{
\subsubsection{Reward Function} 
The reward function is designed to guide the agent toward desired locomotion behaviors, including tracking commanded linear and angular velocities, maintaining body height, and preserving orientation. 
Table \ref{tab:rewards} summarizes the reward terms and their scales across different robots.  
The rewards are grouped into two categories: \textit{Core Locomotion Rewards} encourage stable and efficient motion, \textit{Environment-Constraint Rewards} penalize collisions, stumbles, edge steps, and immobility.

Unlike prior works that manually select waypoints along a preset trajectory to compute tracking rewards, we adopt a simpler formulation from \cite{wmp}:  

{\small \vspace{-5pt}
\begin{equation}
\begin{aligned}    
    r_{\text{lin tracking}} = \exp\Big(-\big( \min ( v_{xy}, v^\mathrm{cmd}_{xy} + 0.1)- v^\mathrm{cmd}_{xy}\big)^2 / \sigma \Big),   
\end{aligned}
\end{equation}}
where $v_{xy}$ is the current planar speed of the robot, $v_{xy}^{\mathrm{cmd}}$ is the commanded speed, and $\sigma$ is a scaling factor. This formulation allows the agent to naturally accelerate, decelerate, or turn as required by the terrain (e.g., when jumping over gaps), reducing human effort in waypoint selection.  
}

\vspace{-2pt}
\begin{table}[htp!]
    \captionsetup{font=footnotesize}
    \caption{Reward scales used for training, grouped into core locomotion and environment-constraint rewards.}
    \centering
    \label{tab:rewards}
    \scriptsize
    \setlength{\tabcolsep}{5pt}
    \setlength{\arrayrulewidth}{1.2pt}
    \renewcommand{\arraystretch}{1.3}
    \begin{tabular}{lccc}
    \hline
     & \textbf{\thead{\scriptsize Go1 \\ \scriptsize (Quadruped)}} & \textbf{\thead{\scriptsize Cassie \\ \scriptsize (Biped)}} & \textbf{\thead{\scriptsize TronA1-W \\ \scriptsize (Wheeled Biped)}} \\
    \hline
    \multicolumn{4}{c}{\textit{Core Locomotion Rewards}} \\
    Linear Velocity Tracking & $1.5$ & $2.5$ & $5.0$ \\
    Angular Velocity Tracking & $0.5$ & $0.5$ & $2.5$ \\
    Torques Penalty & $1\text{e}^{-7}$ & $1\text{e}^{-7}$ & $1\text{e}^{-7}$ \\
    DoF Acceleration Penalty & $2.5\text{e}^{-7}$ & $2.5\text{e}^{-7}$ & $2.5\text{e}^{-7}$ \\
    Action Rate & $-0.03$ & $-0.03$ & $-0.03$ \\
    DoF Error & $-0.04$ & $-0.04$ & $-0.04$ \\
    Z-Linear Velocity Penalty & $-1.0$ & $-1.0$ & $-1.0$ \\
    Feet Air Time & $0.5$ & $0.5$ & $0.5$ \\
    % No Fly & $0.0$ & $0.025$ & $0.0$ \\
    Wheel Penalty & $--$ & $--$ & $-0.01$ \\
    
    \multicolumn{4}{c}{\textit{Environment-Constraint Rewards}} \\
    Collision & $-1.0$ & $-1.0$ & $-1.0$ \\
    Feet Stumble & $-0.1$ & $-0.1$ & $-0.1$ \\
    Feet Edge & $-1.0$ & $-1.0$ & $0.0$ \\
    Cheat & $-1.0$ & $-1.0$ & $-0.7$ \\
    Stuck & $-1.0$ & $-1.0$ & $-1.0$ \\
    \hline
    \end{tabular}
\end{table}
\vspace{-8pt}

\subsection{Model Architecture}

Our training framework consists of an {Internal Model} combined with an {Asymmetric Actor-Critic} setup (Fig. \ref{fig:training}). The Internal Model learns a predictive representation, enabling the policy to reason in a compact state space that is robust to sensor noise and domain shifts.

\subsubsection{Internal Model}
\label{sc:model_architecture}
% The model processes proprioceptive and visual observations into compact representations by rolling out future kinodynamic states over a finite horizon.
{
It serves two purposes simultaneously. During \emph{training}, it provides the expert actor with imagined kinodynamic rollouts that encode future terrain geometry, enabling anticipatory behavior without privileged access to the simulator. 
\footnote{{We adopt the RSSM architecture from Dreamer~\cite{dreamer} as the internal model backbone; we refer to its policy as the \emph{Internal Policy} $\pi_\varphi$ to distinguish it from the \emph{Expert Actor} $\pi_\theta$.}}
During \emph{deployment}, it acts as the dynamics and return oracle for the MPC planner (Algorithm~\ref{alg:mpc}), enabling trajectory optimization in a compact learned state space. This dual role motivates the specific architecture choices below: a gated recurrent unit (GRU) for temporal memory, a stochastic latent state for uncertainty representation, and an explicit kinodynamic CoM state $x$ for interpretable constraint enforcement. 
% The model processes proprioceptive and visual observations into compact representations by rolling out future kinodynamic states over a finite horizon.

The model state is defined as $y = [x, h, z]^\top$, with $h$ representing the recurrent hidden state and $z$ the stochastic latent state, both following the formulation in \cite{dreamerV3}. $x$ is the kinodynamic CoM state from \cite{feedback_mpc} (which includes body pose, twist, and joint states). Explicitly incorporating $x$ ensures interpretability and facilitates easier constraint handling during planning. 
The Internal Model operates at $10$ Hz rather than the actor's $50$ Hz due to the computational cost of depth-image encoding and latent-state estimation; the recurrent hidden state $h$ is thus held fixed between GRU updates and reused by the actor at the higher control frequency.
}

\vspace{-2pt}
\begin{algorithm}[htp!]
    \captionsetup{font=small}
    \caption{Imagination Augmentation}
    \label{alg:internal_model}
    \footnotesize
    \begin{algorithmic}[1]
    \Require $y_0 = [x_0, h_0, z_0]$; observation embedding $\mathbf{e}$; horizon $H$
    \State $\mathcal{X} \gets \emptyset$
    \For{$k = 0$ to $H-1$}
        \State $a_k \sim \pi_\varphi(\ \cdot \mid y_k)$ \hfill \textit{\grey{// Internal Policy}}
        \State $h_{k+1} = g_\varphi(y_k, a_k)$ \hfill \textit{\grey{// GRU update}}
        \If{$k = 0$}
            \State $z_{k+1} \sim p^z_\varphi(\ \cdot \mid \mathbf{e}, h_{k+1})$ \hfill \textit{\grey{// Encoder (posterior)}}
            \State $x_{k+1} = p^x_\varphi(\mathbf{e}, h_{k+1})$ \hfill \textit{\grey{// Estimation}}
        \Else
            \State $z_{k+1} \sim d^z_\varphi(\ \cdot \mid h_{k+1})$ \hfill \textit{\grey{// Latent Dynamics (prior)}}
            \State $x_{k+1} = d^x_\varphi(h_{k+1})$ \hfill \textit{\grey{// Rigid Body Dynamics}}
        \EndIf
        \State $\mathcal{X} \gets \mathcal{X} \cup \{ x_{k+1} \}$
    \EndFor
    \State \Return $\mathcal{X}, y_1 = [x_1, h_1, z_1]$
    \end{algorithmic}
\end{algorithm}
\vspace{-2pt}

The whole model consists of the following components:
\begin{itemize}
    \item \textit{Internal Policy} $\pi_\varphi$: Outputs the action distribution conditioned on the model state $y$.
    \item \textit{GRU Cell} $g_\varphi$: Updates the recurrent memory $h$ for the current model state and action input.
    \item \textit{Encoder} $(p^z_\varphi, p^x_\varphi)$: Consists of two components to estimate the future state variables $(z, x)$ conditioned on the observation embedding $\mathbf{e}$ and the recurrent state $h$. Here $\mathbf{e}$ is obtained by passing $\mathbf{o}$ through a preprocessing module (see Fig. \ref{fig:training}).
    \item \textit{Dynamics} $(d^z_\varphi, d^x_\varphi)$: Predicts future states $(z,x)$ from the recurrent state $h$. It also consists of two components, an MLP $d^z_\varphi$ to model the dynamics probability distribution for the latent state $z$. And $d^x_\varphi$ for the rigid body dynamics equations from \cite{feedback_mpc} to update $x$. This makes the overall dynamics more interpretable and useful for adding constraints during deployment.
    \item \textit{Decoder} $q_\varphi$: Reconstructs the full proprioceptive observations and the depth image given the model state.
    \item \textit{Reward and Value Models} $(r_\varphi, V_\varphi)$: Estimate task reward and value for a given model state.
\end{itemize}

% To note here, $d_\varphi$ consists of two components, an MLP to model a probability distribution for the latent state $z$. The kinodynamic CoM state $x$ is updated using the rigid body dynamics equations from \cite{feedback_mpc}. This makes $d_\varphi$ more interpretable and useful for adding constraints during deployment.

Starting from the {previous model state} $y_0$ and current observation embedding $\mathbf{e}$, Algorithm \ref{alg:internal_model} predicts a horizon $H$ sequence of kinodynamic states entirely in model state. The first step ($k=0$) uses the encoder for a posterior update; subsequent steps use the dynamics prior. Actions are sampled from the Internal Policy at every step. 

% In Algorithm \ref{alg:internal_model}, \textit{Lines 3--4}, the policy outputs actions based on the current model state, and the GRU Cell updates temporal memory.
% In \textit{Lines 6--7}, at $k=0$, the encoder does a posterior update from the observation embedding $\mathbf{e}$ to estimate the new state; and, in \textit{Lines 9--10}, the dynamics does an update for rollout prediction.
% At \textit{Line 12:} the predicted kinodynamic CoM states $\{x_{k+1}\}$ form the imagined rollout $\mathcal{X}$, while the next model state $y_1$ seeds the next cycle.

\subsubsection{Asymmetric Actor-Critic}
The \textit{Expert Actor} $\pi_\theta$ is an MLP that interacts with the environment during training, taking input as $\mathbf{o}$, and $\mathbf{sg}(h, \mathcal{X})$. Here \( \mathbf{sg}(\cdot) \) denotes stop-gradient, preventing Actor-Critic updates from altering the Internal Model.
The \textit{Privileged Critic} $V_\theta$ is trained with access to privileged information $\mathbf{o}^\mathrm{priv}$ and also $\mathbf{sg}(h, \mathcal{X})$.

\subsection{Training}

This section describes the procedure for jointly optimizing the {Internal Model}, {Expert Actor}, and {Privileged Critic}, following the architecture in Section \ref{sc:model_architecture}. 

\subsubsection{Supervised Learning of the Internal Model}
\label{sc:im_loss}

We train it using targets from simulation interaction. For each timestep \( t \) in the replay buffer, we construct: $(\mathbf{o}_t,\ a_t^{\mathrm{expert}},\ r_t^{\mathrm{sim}},\ V_t^{\mathrm{target}},\ x_t^{\mathrm{sim}},\ x_{t+1}^{\mathrm{sim}})$
where \( a_t^{\mathrm{expert}} \) is the action from the Expert Actor, \( r_t^{\mathrm{sim}} \) is the ground-truth reward, \( V_t^{\mathrm{target}} \) is the value target, and \( (x_t^{\mathrm{sim}}, x_{t+1}^{\mathrm{sim}}) \) are ground-truth kinodynamic CoM states.
Recall that the model state is \( y_t = [x_t, h_t, z_t]^\top \), and the model includes a preprocessing module for obtaining $\mathbf{e}_t$ from observations, a GRU core for memory updates, an encoder for posterior estimation, and a dynamics model for prior prediction. 
The decoder, reward, Internal Policy, and Internal Critic networks are modeled as distributions. Thus the combined model loss is defined as:

{\footnotesize\vspace{-10pt}
\begin{equation}\label{eq:im_loss}
\resizebox{\linewidth}{!}{$%
\begin{aligned}
&\mathcal{L}_{\text{IM}} =
\sum_{t=0}^{L} \
\mathbb{E}_{p_\varphi} \Bigg[
    -
    \underbrace{
        \ln r_\varphi(r_t^{\mathrm{sim}} \mid y_t, a_t^{\mathrm{expert}})
    }_{\text{\scriptsize Reward NLL}} -
    \underbrace{
        \ln V_\varphi(V_t^{\mathrm{target}} \mid y_t)
    }_{\text{\scriptsize Value NLL}}
    \\
    &\quad+
    \underbrace{
        \beta \, \mathrm{KL}\Big(p^z_\varphi(z_t \mid \mathbf{e}_t, h_t) \parallel d^z_\varphi(z_t \mid h_t)\Big)
    }_{\text{\scriptsize Latent KL}}
    -
    \underbrace{
        \ln \pi_\varphi(a_t^{\mathrm{expert}} \mid y_t)
    }_{\text{\scriptsize Action Cloning}}     \\ 
    &\quad+
    \underbrace{
        \Big\| x_t^{\mathrm{sim}} - p^x_\varphi(\mathbf{e}_t, h_t) \Big\|_2^2 + \Big\| x_{t+1}^{\mathrm{sim}} - d^x_\varphi(h_t) \Big\|_2^2
    }_{\text{\scriptsize Kinodynamic CoM Loss}} 
    -
    \underbrace{
        \ln q_\varphi(\mathbf{o}_t \mid y_t)
    }_{\text{\scriptsize Reconstruction}}
\Bigg]
\end{aligned}
$}
\end{equation}
}

Here, the KL term aligns the encoder posterior and dynamics prior for consistent imagination; the CoM loss enforces accurate physical state predictions; reward and value terms use negative log-likelihood (NLL) for likelihood maximization under predicted distributions; the behavior cloning term trains the Internal Policy to match expert actions; and the reconstruction term ensures the latent state retains observation information. This supervised approach enables sample-efficient training without propagating gradients into the imagined rollouts from on-policy updates.

{We adopt a variational objective for our internal model rather than a consistency-based one (as in TD-MPC \cite{tdmpc, tdmpc2}) for a specific reason: the privileged critic's scan dot inputs provide sufficient height information for open terrains but cannot capture lateral geometry constraints such as wall clearances that define tasks like crawl and tilt. A consistency loss trained against this privileged representation inherits its blind spots. 
The variational objective instead avoids dependence on the quality of the privileged scan data and produces representations better suited to tasks such as crawl and tilt, as confirmed by the ablation in Section \ref{sc:results_training}.
}

\subsubsection{Policy Learning via PPO}
\label{sc:rl_loss}

The Expert Actor \( \pi_\theta \) and Privileged Critic \( V_\theta \) are trained using PPO on trajectories collected in simulation. As mentioned earlier, the actor receives \( (\mathbf{o}_t, h_t, \mathcal{X}_t) \), where \(\mathcal{X}_t\) is the imagined kinodynamic rollout from Algorithm \ref{alg:internal_model}. The critic additionally uses privileged simulator information. The PPO objective in our approach is defined as:

{\footnotesize\vspace{-10pt}
\begin{equation}
\resizebox{\linewidth}{!}{$%
\begin{aligned}
&\mathcal{L}_\text{PPO} = \mathbb{E}\Bigg[
\underbrace{\min\Big(\rho_t(\theta)\hat{A}_t,\ \mathrm{clip}\big(\rho_t(\theta),1-\epsilon,1+\epsilon\big)\hat{A}_t\Big)}_{\text{\scriptsize Clipped Policy Loss}} \\
&\quad + \underbrace{\beta_{\mathrm{ent}}\mathcal{H}\Big(\pi_\theta(\ \cdot \mid \mathbf{o}_t,\mathbf{sg}(h_t,\mathcal{X}_t))\Big)}_{\text{\scriptsize Entropy Bonus}}
- \underbrace{\Big\| V_\theta(\mathbf{o}_t^{\mathrm{priv}},\mathbf{sg}(h_t,\mathcal{X}_t)) - \hat{R}_t \Big\|^2_2}_{\text{\scriptsize Value Loss}}
\Bigg]
\end{aligned}
$}
\end{equation}
}

Here, \( \rho_t(\theta) \) is the probability ratio between new and old policies, \( \hat{A}_t \) is the GAE advantage estimate, and \( \hat{R}_t \) is the return target.

\subsection{Deployment: Infinite-Horizon Planning}
\label{sc:planning}

Safe and optimal locomotion requires online filtering of actions generated by the learned policy (in our case $\pi_\theta$). While RL policies provide generalization, they may produce unsafe or suboptimal actions under distribution shifts or unseen situations. An online planner is therefore essential to ensure stability, safety, and constraint satisfaction during deployment. Furthermore, incorporating model-based planning improves interpretability by explicitly reasoning over predicted kinodynamic states.

To this end, we adopt a planning approach using the learned models from Section \ref{sc:model_architecture}. The planner optimizes a receding-horizon trajectory with value estimates for bootstrapping to approximate infinite-horizon return. At each control step starting from $y_0$, the objective is to compute an optimal action sequence \(a^*_{0:H-1}\) as formulated below,

{\small \vspace{-5pt}
\begin{equation}\label{eq:inf_mpc}
\begin{aligned}
\max_{a_{0:H-1}} \ \ \ &
\mathbb{E}\Bigg[ \sum_{k=0}^{H-1} \gamma^k r_k + \gamma^H V_H \Bigg] \\
\text{s.t. } \; &
y_{k+1} \sim f_\varphi(\ \cdot \mid y_{k}, a_{k}) \ \ \forall k = 0,\dots,H-1, \\
& r_k \sim r_\varphi(\ \cdot \mid y_{k}, a_{k}) \ \ \forall k = 0,\dots,H-1, \\
& V_H \sim V_\varphi(\ \cdot \mid y_{H}), \\
& c_j(y_{k}, a_{k}) \le 0 \ \ \forall j \in \mathcal{C}, \ \ \ a_{k} \in \mathcal{A}.
\end{aligned}
\end{equation}
}

Here, \(f_\varphi(\cdot)\) denotes the composition of $(d^z_\varphi, d^x_\varphi)$ and $g_\varphi$ for readability. 
{The constraint set $\mathcal{C}$ enforces kinodynamic feasibility and safety, including: (i) joint position and velocity limits, (ii) base twist bounds on linear and angular velocity}. And \(\mathcal{A}\) represents an admissible set for actions.
% The constraint set \(\mathcal{C}\) enforces kinodynamic feasibility and safety (e.g., twist or joint limits), while \(\mathcal{A}\) represents an admissible set for actions.
Since Equation \eqref{eq:inf_mpc} is a nonlinear stochastic program, exact solutions are intractable for real-time execution. We adopt a constrained version of the Model Predictive Path Integral (MPPI) \cite{mppi} method (see Algorithm \ref{alg:mpc}), which updates a Gaussian distribution over action sequences using elite sampling.
Here, $\mu$ and $\sigma$ denote the mean and covariance of this Gaussian policy with RL policy providing a warm-start to accelerate convergence. 
% The algorithm operates in a receding-horizon manner, applying the first optimized action at each step.

The infinite-horizon objective in Equation \eqref{eq:inf_mpc} uses value function bootstrapping at the terminal state to account for rewards beyond the planning horizon. This improves long-term planning capabilities compared to classical fixed-horizon MPC formulations \cite{cvx_mpc, jumpmpc}. Furthermore, although we adopted the MPPI formulation for computational efficiency, our framework is general enough to employ other gradient or Hessian based optimization solvers.

\begin{algorithm}[htp!]
\footnotesize
\captionsetup{font=small}
\caption{Data-Driven $\infty$ Horizon MPC (Deployment)}
\label{alg:mpc}
\begin{algorithmic}[1]
\Require 
$(\mu_{\mathrm{prev}}, \sigma_{\mathrm{prev}})$: Previous Gaussian policy; 
$y_{\mathrm{prev}}$: Previous model state; 
$\mathbf{e}$: Observation embedding; 
$H$: Horizon;
$N$: \# of MPPI updates; 
$M$: \# of MPPI samples; 
$M_\pi$: \# of RL policy samples; 
$M_\mathrm{elite}$: Elite set size; 
$\alpha$: Temporal momentum; $\beta$: Iteration momentum.

\State Get $y_0$ and a Gaussian policy $(\mu_{\mathrm{RL}}, \sigma_{\mathrm{RL}})$ using expert actor $\pi_\theta$ starting from $(\mathbf{e}, y_\mathrm{prev})$ up to horizon $H$. \hfill \textit{\grey{// Uses Algorithm \ref{alg:internal_model}}}
\State $(\mu_0, \sigma_0) \gets \alpha (\mu_{\mathrm{prev}}, \sigma_{\mathrm{prev}}) + (1 - \alpha) (\mu_\mathrm{RL}, \sigma_\mathrm{RL})$

\For{$i = 1 \ \text{to} \ N$} 
    \State $A \gets \emptyset$, $R \gets \emptyset$

    \State \textit{{\grey{// Sample action trajectories}}}
    \[
    a^{(j)}_{0:H-1} \sim \mathcal{N}(\mu_{i-1}, \sigma_{i-1}), \quad \forall j \in [1,M]
    \]
    \[
    a^{(j)}_{0:H-1} \sim \mathcal{N}(\mu_\mathrm{RL}, \sigma_\mathrm{RL}), \quad j \in [M + 1, M + M_\pi]
    \]

    \State \textit{{\grey{// Evaluate expected return of trajectories}}} \footnotemark
    \[
    \bar{R}^{(j)} = 
    \mathbb{E}\left[
        \sum_{k=0}^{H-1} \gamma^k r^{(j)}_k + \gamma^H V^{(j)}_H
    \right]
    \]
    \State $R \gets R \cup \{\bar{R}^{(j)}\}$

    \State Sort $A$ according to returns $R$
    
    \State $A_E \gets$ select top-$M_{\mathrm{elite}}$ trajectories from $A$ satisfying $\mathcal{C}$
    
    \State $(\mu_{\mathrm{elite}}, \sigma_{\mathrm{elite}}) \gets$ fit $\mathcal{N}$ to the elite set $A_E$  \hfill \textit{\grey{// MPPI Update}}
    
    \State $(\mu_i, \sigma_i) \gets \beta (\mu_{\mathrm{elite}}, \sigma_{\mathrm{elite}}) + (1 - \beta) (\mu_{i-1}, \sigma_{i-1})$
\EndFor

\State \Return First action $a_0 \sim \mathcal{N}(\mu_N, \sigma_N)$
\end{algorithmic}
\end{algorithm}

\footnotetext{The expectation is taken over trajectories induced by $y^{(j)}_{k+1} \sim f_\varphi(\ \cdot \mid y^{(j)}_k, a^{(j)}_k), r^{(j)}_k \sim r_\varphi(\ \cdot \mid y^{(j)}_k, a^{(j)}_k)$ and $V^{(j)}_H \sim V_\varphi(\ \cdot \mid y^{(j)}_H)$.}

\section{Results}\label{sc:results}

We evaluate VIP-Loco through extensive simulation experiments, complemented by ablation studies to analyze the contribution of each framework component. Training was performed in an open-source environment built on NVIDIA's Isaac Gym \cite{isaac_gym, leggedgym}, with all neural networks implemented in PyTorch \cite{pytorch}. Robots were trained to navigate diverse terrains, including rough slopes, staircases, gaps requiring jumps, obstacles for climbing, narrow passages for tilting, and low-clearance slabs for crawling. To enhance robustness, we employed domain randomization and a curriculum learning strategy that gradually increased terrain difficulty.
% Table \ref{tab:terrain} outlines the terrain curriculum for the quadruped; parameters for the biped and wheeled biped were scaled according to their respective dimensions. 
{Each morphology is trained with an independent model using robot-specific reward scales (Table~\ref{tab:rewards}) and terrain curricula. We demonstrate that the VIP-Loco framework architecture is sufficiently general to be instantiated across fundamentally different morphologies without requiring structural changes to the method.}

% \begin{table}[htp!]
%     \scriptsize
%     \centering
%     \captionsetup{font=footnotesize}
%     \caption{Terrain level curriculum for Go1 (Quadruped). Cassie (Biped) and TronA1-W (Wheeled Biped) use proportionally scaled values.}
%     \setlength{\tabcolsep}{5pt}
%     \setlength{\arrayrulewidth}{1.5pt}
%     \renewcommand{\arraystretch}{1.3}
%     \begin{tabular}{c | c c c c c c}
%         \hline
%         \textbf{\thead{\scriptsize Level}} & \textbf{\thead{\scriptsize Slope \\ \scriptsize Range}} & \textbf{\thead{\scriptsize Stair \\ \scriptsize Height}} & \textbf{\thead{\scriptsize Climb \\ \scriptsize Height}} & \textbf{\thead{\scriptsize Crawl \\ \scriptsize Height}} & \textbf{\thead{\scriptsize Tilt \\ \scriptsize Width}} & \textbf{\thead{\scriptsize Gap \\ \scriptsize Length}} \\
%         \hline
%         0 (min) & $\pm$5$^\circ$  & 5 cm  & 10 cm & 40 cm & 40 cm & 5 cm \\
%         8 (max) & $\pm$30$^\circ$ & 16 cm & 54 cm & 21 cm & 28 cm & 90 cm \\
%         \hline
%     \end{tabular}
%     \label{tab:terrain}
%     \vspace{-5pt}
% \end{table}

\begin{figure}[htp!]
    \centering
    \scriptsize
    \captionsetup{font=footnotesize}
    \includegraphics[width=0.75\linewidth]{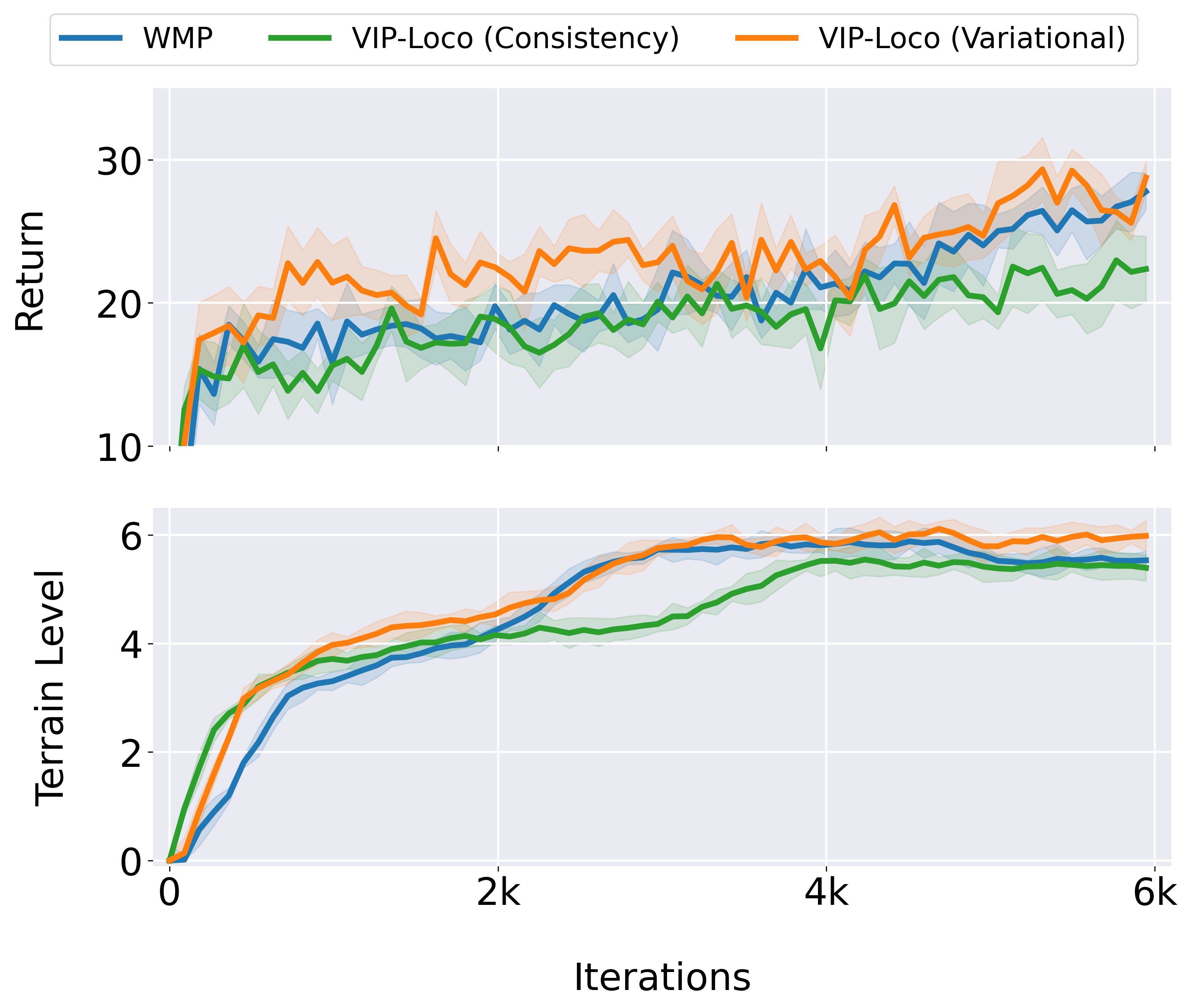}
    \caption{\textit{Training comparison for Go1 (quadruped) over $5$ seeds --} 
    Top: episodic return progression over training iterations. Bottom: average terrain level successfully mastered. VIP-Loco (Variational) achieves the highest and most stable returns, steadily mastering harder terrains (levels $\ge$ 6). WMP performs competitively but asymptotically regresses to easier terrains. VIP-Loco (Consistency) converges earlier and stagnates at lower levels.
    % , highlighting the benefits of the variational loss for robust representations.
    }
    \label{fig:ablation_plot}
    \vspace{-5pt}
\end{figure}

\begin{figure*}[htp!]
    \vspace*{0.2cm}
    \scriptsize
    \centering
    \captionsetup{font=footnotesize}
    \includegraphics[width=0.7\linewidth]{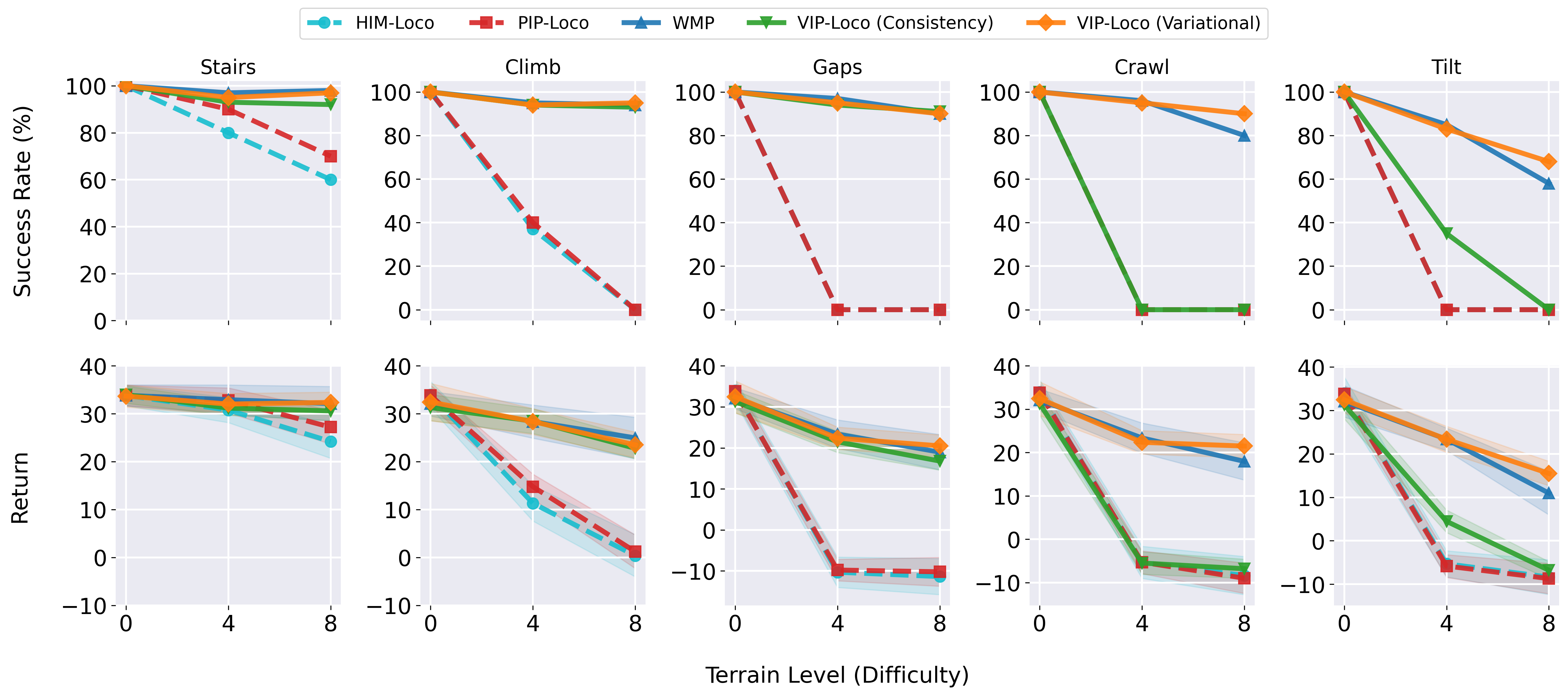}
    \caption{\textit{Comparison of locomotion performance for Go1 (quadruped) across terrains of increasing difficulty and $5$ seeds --}
     The plots show the success rate (top row) and average return (bottom row) for five locomotion methods: HIM-Loco, PIP-Loco, WMP, VIP-Loco (with Consistency Loss), and VIP-Loco (with Variational Loss). Each column corresponds to a different terrain type and the x-axis indicates terrain difficulty level (0 = easiest, 8 = hardest).
    }
    \label{fig:comparison}
    \vspace{-5pt}
\end{figure*}

\begin{table*}[htp!]
    \scriptsize
    \centering
    \captionsetup{font=footnotesize}
    \caption{Average episodic return comparison across six terrains and three robots after $5$k iterations (averaged over $5$ seeds). 
    % The \% Gain column reports the relative improvement of VIP-Loco over the baseline without planning.
    }

    \setlength{\tabcolsep}{4pt}
    \setlength{\arrayrulewidth}{1.5pt}
    \renewcommand{\arraystretch}{1.3}
    \begin{tabular}{c | c | c c c c c c }
    \hline
    \textbf{Robot} & \textbf{Method} & \textbf{Slopes} & \textbf{Stairs} & \textbf{Climb} & \textbf{Gaps} & \textbf{Crawl} & \textbf{Tilt} \\
    \hline
    \multirow{2}{*}{\thead{\scriptsize Go1 \\ \scriptsize (Quadruped)}} 
    & VIP-Loco w/o Planning & $\mathbf{28.37 \pm 2.59}$ & $26.66 \pm 2.54$          & $22.80 \pm 3.29$          & $24.80 \pm 3.29$          & $20.80 \pm 2.82$          & $16.18 \pm 2.54$          \\
    & VIP-Loco              & ${27.22 \pm 1.34}$        & $\mathbf{27.85 \pm 3.28}$ & $\mathbf{23.26 \pm 2.45}$ & $\mathbf{26.33 \pm 2.38}$ & $\mathbf{22.63 \pm 3.74}$ & $\mathbf{18.78 \pm 2.13}$ \\
    \hline
    \multirow{2}{*}{\thead{\scriptsize Cassie \\ \scriptsize (Biped)}}
    & VIP-Loco w/o Planning & ${22.54 \pm 1.83}$        & $23.66 \pm 3.79$          & $21.46 \pm 2.46$          & $\mathbf{21.34 \pm 2.91}$ & $\mathbf{18.52 \pm 1.91}$ & $12.33 \pm 1.98$          \\
    & VIP-Loco              & $\mathbf{23.19 \pm 2.94}$ & $\mathbf{23.85 \pm 1.64}$ & $\mathbf{27.42 \pm 2.72}$ & $20.50 \pm 3.06$          & $17.69 \pm 2.22$          & $\mathbf{14.40 \pm 3.21}$ \\
    \hline
    \multirow{2}{*}{\thead{\scriptsize TronA1-W \\ \scriptsize (Wheeled Biped)}}
    & VIP-Loco w/o Planning & $33.47 \pm 1.38$          & $31.32 \pm 1.22$          & $29.39 \pm 3.92$          & $\mathbf{31.19 \pm 2.26}$ & $27.55 \pm 2.73$          & $21.82 \pm 2.01$          \\
    & VIP-Loco              & $\mathbf{38.29 \pm 2.39}$ & $\mathbf{34.05 \pm 2.31}$ & $\mathbf{32.87 \pm 1.50}$ & $30.22 \pm 2.99$          & $\mathbf{29.27 \pm 3.09}$ & $\mathbf{24.03 \pm 1.17}$ \\
    \hline
    \end{tabular}
    \label{tab:combined_robot_results}
    \vspace{-5pt}
\end{table*}

We train using PPO \cite{ppo} with a clipping range of 0.2, a GAE factor of 0.95, and a discount factor of 0.99. All networks are optimized using Adam with a learning rate of 0.001. Training was executed on a desktop equipped with an Intel Xeon Gold 5318Y (48) @ 3.40 GHz and two NVIDIA RTX A6000 GPUs. 
Our implementation also uses the NVIDIA WARP library to maintain VRAM usage below 25 GB while supporting 4096 parallel environments and rendering depth images within milliseconds.
% For deployment, we run the planner (Algorithm \ref{alg:mpc}) using JAX \cite{jax}, achieving inference rates of 40-50 Hz.
{For deployment, we run the planner (Algorithm~\ref{alg:mpc}) using JAX~\cite{jax}, achieving inference rates of 40--50~Hz. An approximate per-component breakdown at deployment is as follows: depth image encoding ($\sim$2--3~ms), GRU state update ($<$1~ms), imagination rollout over horizon $H$ ($\sim$3--5~ms), MPPI sampling and return evaluation over $M$ trajectories ($\sim$10--15~ms), and elite selection and distribution update ($<$1~ms), yielding a total planner latency of approximately 20--25~ms per control step on a desktop GPU.
}

\subsubsection{Ablation Methods}

{
To evaluate VIP-Loco systematically, we compare it against the following methods, which together isolate the contributions of proprioception, vision, and planning:

\begin{itemize}
    \item \textbf{HIM-Loco}~\cite{himloco}: A proprioception-only baseline using internal model latent representations, without vision or planning. 
    % Establishes the floor for what structured state prediction alone achieves.

    \item \textbf{PIP-Loco}~\cite{piploco}: Extends HIM-Loco with a hierarchical internal model and infinite-horizon MPC at deployment, but uses only proprioceptive inputs. Isolates the effect of planning \emph{without} vision.

    \item \textbf{WMP}~\cite{wmp}: A visually-guided locomotion method using a world model with depth inputs and RL, but \emph{without} explicit MPC-based planning at deployment. Isolates the effect of vision \emph{without} planning.

    \item \textbf{VIP-Loco (Consistency)}: A variant of our method that replaces the variational latent loss with the latent consistency loss from~\cite{tdmpc,tdmpc2}. 
    Isolates the effect of the representation learning objective.

    \item \textbf{VIP-Loco (Variational)}: Our proposed method, trained with the variational loss in Eq.~(\ref{eq:im_loss}), and infinite-horizon MPC at deployment.
\end{itemize}

Together, HIM-Loco and PIP-Loco represent the [RL + MPC] family (no vision); WMP represents [RL + Vision] (no planning); and VIP-Loco represents [RL + Vision + MPC], enabling analysis of each component's contribution.}

\subsection{Training Comparison}
\label{sc:results_training}

Figure \ref{fig:ablation_plot} shows the training performance of WMP, VIP-Loco (Consistency), and VIP-Loco (Variational) over $6000$ iterations, tracking average episodic return (top) and mastered terrain level (bottom). All methods adapt quickly, showing rapid early gains in both return and terrain level. However, their asymptotic behavior diverges: VIP-Loco (Variational) achieves the highest and most stable returns while steadily mastering harder terrains, sustaining levels above 6. WMP performs competitively and it reaches similar terrain levels, but it asymptotically regresses to easier terrains to maintain higher reward. 
% In contrast, VIP-Loco (Consistency) plateaus early, converging to lower returns and stagnating near level 5, struggling with challenging terrains like crawl and tilt, likely due to representation not being able to use the limited information in the privileged critic’s scan dot inputs. 
In contrast, VIP-Loco (Consistency) plateaus at a lower level, struggling on terrains like crawl and tilt. 
% This suggests its loss function fails to learn effective representations when the privileged critic's sparse scan data provides an insufficient description of the terrain.
% Overall, while all methods show fast initial adaptation, only VIP-Loco (Variational) and, to some extent, WMP maintain strong returns and terrain generalization throughout training.

{
Overall, while all methods show fast initial adaptation, only VIP-Loco (Variational) and, to some extent, WMP maintain strong returns and terrain generalization throughout training. We hypothesize that the failure of the consistency-based variant on crawl and tilt stems from an information asymmetry: the privileged critic's scan dot inputs provide sufficient height information for open terrains but cannot capture the lateral geometry constraints (e.g., wall clearances) that define these tasks, causing the consistency loss to learn representations poorly suited to visually demanding locomotion.
The variational objective, by directly reconstructing the depth image, learns more effective representations when sparse scan data provides an insufficient description of the terrain.
}

\begin{figure*}[htp!]
    \vspace*{0.2cm}

    \centering
    \scriptsize
    \captionsetup{font=footnotesize}
    % First subfigure
    \begin{subfigure}[b]{0.9\linewidth}
        \centering
        \includegraphics[width=0.9\linewidth]{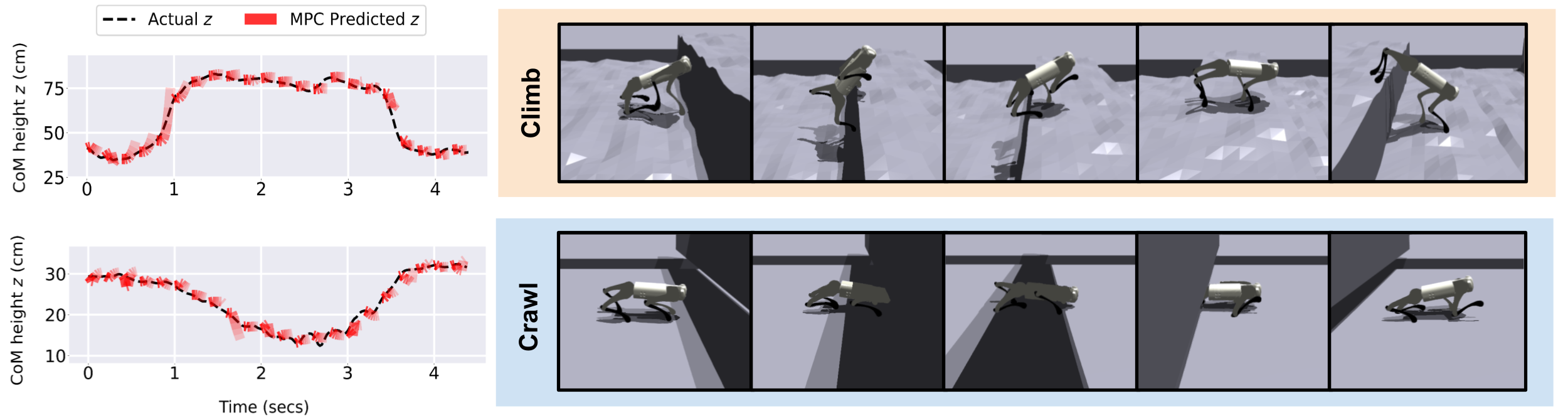}        
        \caption{Go1 (quadruped) performing \emph{Climb} and \emph{Crawl}. The left plots show predicted vs. actual CoM height, while the right frames show execution snapshots demonstrating coordinated four-leg motion across terrains.}
    \end{subfigure}
    \hfill
    % Second subfigure
    \begin{subfigure}[b]{0.9\linewidth}
        \centering
        \includegraphics[width=0.9\linewidth]{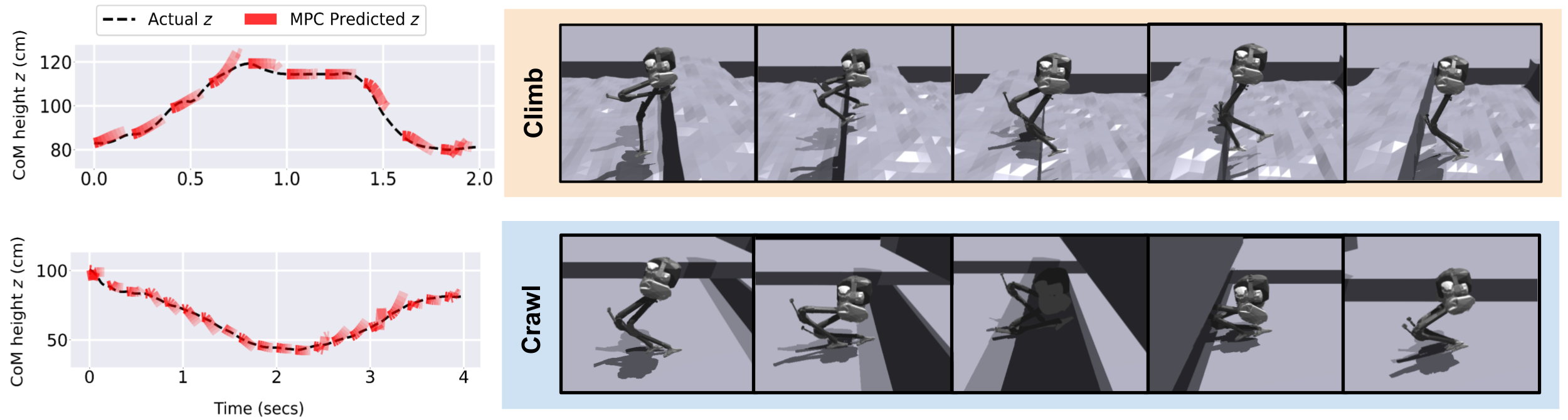}        
        \caption{Cassie (biped) executing the same tasks. The model accurately forecasts sharp height variations during climbing and low crouched profiles during crawling, with snapshots showing balanced bipedal motion.}
    \end{subfigure}
    \hfill
    % Third subfigure
    \begin{subfigure}[b]{0.9\linewidth}
        \centering
        \includegraphics[width=0.9\linewidth]{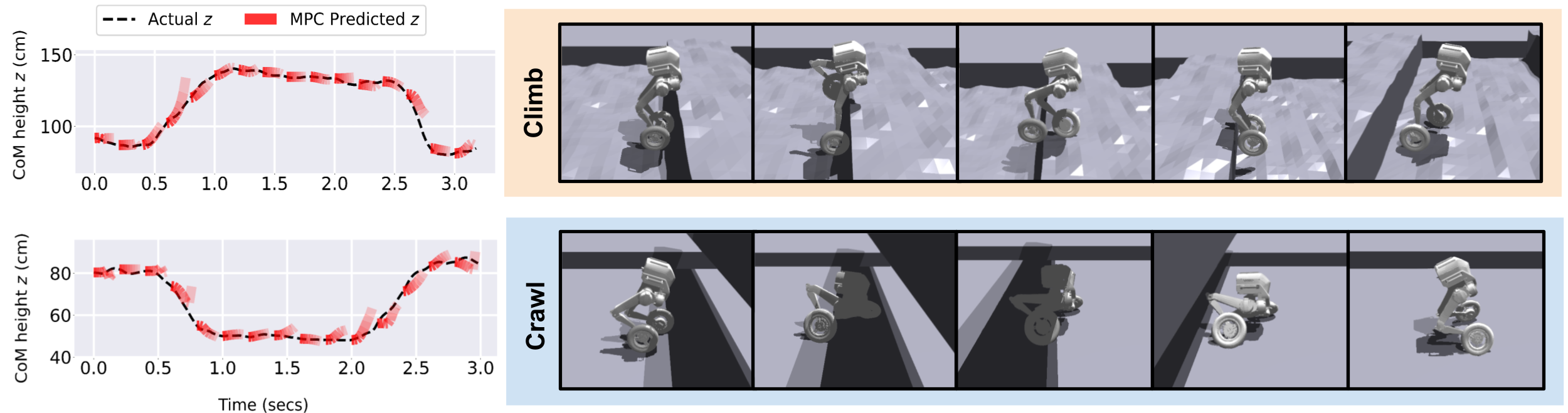}
        \caption{TronA1-W (wheeled biped) combining leg and wheel motions. Predicted and actual profiles align well, and snapshots illustrate seamless hybrid transitions for climbing and crawling.}
    \end{subfigure}

    \caption{\textit{Qualitative evaluation for VIP-Loco with planning across three robot morphologies: (a) Go1 (quadruped), (b) Cassie (biped), and (c) TronA1-W (wheeled biped).} Each subfigure includes (left) predicted vs. actual CoM height trajectories for \emph{Climb} and \emph{Crawl} tasks, and (right) corresponding execution frames. The MPC predictions and actual measurements demonstrate interpretable dynamics modeling, while the snapshots show task-specific stable behaviors across diverse morphologies.}
    \label{fig:qualitative_rollouts}
    \vspace{-5pt}
\end{figure*}

\subsection{Deployment Performance Comparison}

Figure \ref{fig:comparison} compares success rate and episodic return across five of the six terrain types. 
% \textit{Stairs:} All methods succeed on easy levels, but PIP-Loco and HIM-Loco degrade sharply, while VIP-Loco variants maintain $>90\%$ success and high returns even at maximum difficulty. WMP performs moderately, trailing VIP-Loco. 
% \textit{Climb \& Gaps:} PIP-Loco and HIM-Loco fail beyond level 4; WMP and VIP-Loco again achieve near-perfect success and returns throughout, emphasizing the importance of visual modalities.
% \textit{Crawl \& Tilt:} Most methods fail catastrophically beyond level 4; restricted motion space severely impacts all except WMP and VIP-Loco (Variational), which maintain $>60\%$ success rate and strong positive returns.
% Consistency-based VIP-Loco degrades at higher levels. 
% This again highlights that the variational formulation exhibits the highest robustness and generalization when no information can be extracted from scan dot inputs that are sent to the privileged critic.
{On stairs, VIP-Loco variants maintain $>90\%$ success at maximum difficulty while PIP-Loco and HIM-Loco degrade sharply. On visually demanding tasks (climb, gaps), proprioceptive-only methods fail beyond level 4, while vision-based methods remain robust. On crawl and tilt, only WMP and VIP-Loco (Variational) sustain $>60\%$ success, with the consistency variant degrading at higher levels.
This again highlights that the variational formulation exhibits the highest robustness and generalization when no information can be extracted from the scan-dot inputs fed to the privileged critic. We note that VIP-Loco (Variational) and WMP exhibit similar success rates across several terrains (Fig.~\ref{fig:comparison}). The primary advantage of VIP-Loco over WMP lies not in raw success rate but in interpretability and constraint satisfaction enabled by explicit MPC planning, as evidenced by the following section.}

\subsection{Why Combine MPC and RL?}

Table \ref{tab:combined_robot_results} compares VIP-Loco with and without planning across three robots, evaluated over $5000$ iterations on six terrain types and averaged across five seeds. 
While locomotion behaviors appear visually similar, 
% planning consistently yields higher returns across all robots. 
{planning achieves higher average returns across robots, with gains concentrated on anticipatory terrains, though marginal regressions appear on some easier terrain types where the no-planning policy already saturates.}
The performance gains are particularly evident on terrains where stability and precise foothold selection are critical for tasks that benefit from planning’s ability to anticipate future states. 
Among the robots, \emph{TronA1-W (wheeled-biped)} shows the most significant improvement, likely because its hybrid wheeled-biped morphology introduces additional non-holonomic constraints, making trajectory planning in model state $y$ more advantageous.
% As discussed earlier, these improvements are not only limited to performance, but planning also enhances interpretability by providing insight into the robot’s decision-making process and introduces a corrective mechanism that filters and refines the actions proposed by the RL agent. 

{
To make the role of each component explicit: {PIP-Loco}, which represents [RL + MPC] without vision, succeeds on proprioceptive terrains (slopes, stairs) but degrades sharply on visually demanding tasks (gaps, crawl, tilt) beyond difficulty level 4 (Fig.~\ref{fig:comparison}). {WMP} represents [RL + Vision] without planning; it captures terrain geometry effectively but lacks the corrective horizon that filters unsafe or suboptimal actions at deployment. {VIP-Loco} combines both: Table~\ref{tab:combined_robot_results} shows that the planning stage contributes a return gain over the no-planning variant for the quadruped, biped, and wheeled-biped, with the largest gains on terrains where anticipatory foothold selection is critical (gaps, crawl, climb). This confirms that neither vision nor planning alone is sufficient; their combination is what enables robust performance at the highest difficulty levels.
}

% Table \ref{tab:combined_robot_results} presents a comparison of VIP-Loco with and without planning across three robotic platforms, evaluated over $5000$ iterations on six terrain types and averaged across five random seeds. 
% Incorporating planning consistently yields higher returns for all robots, despite the locomotion behaviors appearing visually similar.
% As discussed earlier, in addition to the observed performance improvements, planning enhances interpretability by providing insight into the robot's decision-making process.
% Furthermore, it serves as a corrective mechanism to filter and refine the actions proposed by the reinforcement learning (RL) agent.
% \textit{Go1:} Both versions perform well on smooth terrains (e.g., {Slopes}), where reactive control suffices. However, planning yields better gains on {Stairs}, {Climb}, and {Tilt}, improving returns by $2.5+$ units on the latter, where foresight aids stability. 
% \textit{Cassie:} Planning offers large benefits on terrains requiring vertical maneuvers, e.g., {Climb} ($+6$ units) and {Tilt}, while the non-planning variant performs slightly better on reactive-demanding tasks ({Gaps}, {Crawl}). Overall, planning improves balance-critical behaviors. 
% \textit{TronA1-W.} The wheeled biped gains most from planning, especially on {Slopes}, \emph{Stairs}, and \emph{Climb} ($+3$–$5$ units). Only \emph{Gaps} favors reactive control, possibly due to fast wheel responses. 
% Especially on terrains requiring long-horizon reasoning, while reactive strategies suffice for simpler or highly constrained tasks.

\subsubsection{Qualitative Rollouts and Model Accuracy}

Figure \ref{fig:qualitative_rollouts} shows \emph{Climb} and \emph{Crawl} rollouts for all robots, comparing MPC-predicted CoM height (red) with actual tracked profiles (black).
Predictions align with execution, confirming that VIP-Loco’s learned dynamics model provides insight into the receding-horizon planning. 
Snapshots illustrate morphology-specific behaviors: quadrupeds coordinate legs for climbing/crawling, bipeds adapt step height and posture, and the wheeled biped blends wheel-leg transitions.

\section{Conclusion}

{
We introduced VIP-Loco, a perceptive planning framework for legged locomotion that bridges vision-based RL and infinite-horizon MPC. Our internal model provides compact kinodynamic features enabling imagination augmentation during training and constraint-aware planning at deployment. Experiments across three morphologies demonstrate consistent benefits: 
VIP-Loco yields higher average returns across robots, with the most consistent gains on tasks requiring anticipatory foothold selection, particularly climbing and tilt. Marginal regressions on some terrains (e.g., gaps and crawl for Cassie) indicate that planning benefits are morphology and terrain-dependent.
While results are simulation-based, the framework incorporates domain randomization and deploys at 40–50 Hz, compatible with onboard control loops.
PIP-Loco \cite{piploco}, the closest predecessor of this work, has been validated on physical hardware, providing evidence that the on-policy training plus MPC deployment paradigm transfers to the real world. Extending VIP-Loco to hardware, including depth sensor noise and onboard compute constraints, is the primary direction for future work.

% While all results reported here are in simulation, several design choices directly target sim-to-real transfer. Domain randomization over mass, friction, and motor parameters is applied throughout training, and the asymmetric actor-critic architecture avoids privileged inputs at deployment. The planner operates at 40--50~Hz, a rate compatible with onboard control loops used in prior hardware deployments on Go1 and Cassie-class platforms. The underlying proprioceptive framework, PIP-Loco~\cite{piploco}, has been validated on physical hardware, providing evidence that the on-policy training plus MPC deployment paradigm transfers to the real world. Extending VIP-Loco to hardware including addressing depth sensor noise and onboard compute constraints is a key direction for future work.
}
% By unifying perception, long-horizon reasoning, and real-time feasibility, VIP-Loco represents a step toward robust, adaptive, and safe locomotion in complex real-world environments.
% Future work will try to incorporate more visual modalities and aim to learn a single generalist agent for locomotion as well as manipulation tasks.

\bibliographystyle{ieeetr}
\footnotesize
\bibliography{references}

@article{wmp, 
    title={World Model-based Perception for Visual Legged Locomotion}, 
    author={Hang Lai and Jiahang Cao and Jiafeng Xu and Hongtao Wu and Yunfeng Lin and Tao Kong and Yong Yu and Weinan Zhang},
    journal={arXiv preprint arXiv:2409.16784},
    year={2024}
}

@misc{pie,
      title={PIE: Parkour with Implicit-Explicit Learning Framework for Legged Robots}, 
      author={Shixin Luo and Songbo Li and Ruiqi Yu and Zhicheng Wang and Jun Wu and Qiuguo Zhu},
      year={2024},
      eprint={2408.13740},
      archivePrefix={arXiv},
      primaryClass={cs.RO},
      url={https://arxiv.org/abs/2408.13740}, 
}

@article{dreamwaq++,
      title={Obstacle-Aware Quadrupedal Locomotion With Resilient Multi-Modal Reinforcement Learning},
      author={Nahrendra, I and Yu, Byeongho and Oh, Minho and Lee, Dongkyu and Lee, Seunghyun and Lee, Hyeonwoo and Lim, Hyungtae and Myung, Hyun},
      journal={arXiv preprint arXiv:2409.19709},
      year={2024}
    }

@article{dreamerV3,
  title={Mastering Diverse Domains through World Models},
  author={Hafner, Danijar and Pasukonis, Jurgis and Ba, Jimmy and Lillicrap, Timothy},
  journal={arXiv preprint arXiv:2301.04104},
  year={2023}
}

@misc{multi_brain,
      title={MBC: Multi-Brain Collaborative Control for Quadruped Robots}, 
      author={Hang Liu and Yi Cheng and Rankun Li and Xiaowen Hu and Linqi Ye and Houde Liu},
      year={2024},
      eprint={2409.16460},
      archivePrefix={arXiv},
      primaryClass={cs.RO},
      url={https://arxiv.org/abs/2409.16460}, 
}

@INPROCEEDINGS{piploco,
  author={Shirwatkar, Aditya and Saxena, Naman and Chandra, Kishore and Kolathaya, Shishir},
  booktitle={2025 IEEE International Conference on Robotics and Automation (ICRA)}, 
  title={PIP-Loco: A Proprioceptive Infinite Horizon Planning Framework for Quadrupedal Robot Locomotion}, 
  year={2025},
  volume={},
  number={},
  pages={11198-11204},
  doi={10.1109/ICRA55743.2025.11128382}}

@INPROCEEDINGS{feedback_mpc,
  author={Grandia, Ruben and Farshidian, Farbod and Ranftl, René and Hutter, Marco},
  booktitle={2019 IEEE/RSJ International Conference on Intelligent Robots and Systems (IROS)}, 
  title={Feedback MPC for Torque-Controlled Legged Robots}, 
  year={2019},
  volume={},
  number={},
  pages={4730-4737},
  doi={10.1109/IROS40897.2019.8968251}}

@inproceedings{cvx_mpc,
  title={Dynamic locomotion in the mit cheetah 3 through convex model-predictive control},
  author={Di Carlo, Jared and Wensing, Patrick M and Katz, Benjamin and Bledt, Gerardo and Kim, Sangbae},
  booktitle={2018 IEEE/RSJ international conference on intelligent robots and systems (IROS)},
  pages={1--9},
  year={2018},
  organization={IEEE}
}

@INPROCEEDINGS{nmpc,
  author={Ding, Yanran and Pandala, Abhishek and Park, Hae-Won},
  booktitle={2019 International Conference on Robotics and Automation (ICRA)}, 
  title={Real-time Model Predictive Control for Versatile Dynamic Motions in Quadrupedal Robots}, 
  year={2019},
  volume={},
  number={},
  pages={8484-8490},
  doi={10.1109/ICRA.2019.8793669}}

@article{tdmpc,
  title={Temporal difference learning for model predictive control},
  author={Hansen, Nicklas and Wang, Xiaolong and Su, Hao},
  journal={arXiv preprint arXiv:2203.04955},
  year={2022}
}

@inproceedings{loop,
  title={Learning off-policy with online planning},
  author={Sikchi, Harshit and Zhou, Wenxuan and Held, David},
  booktitle={Conference on Robot Learning},
  pages={1622--1633},
  year={2022},
  organization={PMLR}
}

@inproceedings{leggedgym,
  title={Learning to walk in minutes using massively parallel deep reinforcement learning},
  author={Rudin, Nikita and Hoeller, David and Reist, Philipp and Hutter, Marco},
  booktitle={Conference on Robot Learning},
  pages={91--100},
  year={2022},
  organization={PMLR}
}

@article{rma,
  title={RMA: Rapid Motor Adaptation for Legged Robots},
  author={Kumar, Ashish and Fu, Zipeng and Pathak, Deepak and Malik, Jitendra},
  journal={Robotics: Science and Systems XVII},
  year={2021},
  publisher={Robotics: Science and Systems Foundation}
}

@inproceedings{dreamwaq,
  title={DreamWaQ: Learning robust quadrupedal locomotion with implicit terrain imagination via deep reinforcement learning},
  author={Nahrendra, I Made Aswin and Yu, Byeongho and Myung, Hyun},
  booktitle={2023 IEEE International Conference on Robotics and Automation (ICRA)},
  pages={5078--5084},
  year={2023},
  organization={IEEE}
}

@inproceedings{himloco,
  title={Hybrid internal model: Learning agile legged locomotion with simulated robot response},
  author={Long, Junfeng and Wang, Zirui and Li, Quanyi and Cao, Liu and Gao, Jiawei and Pang, Jiangmiao},
  booktitle={The Twelfth International Conference on Learning Representations},
  year={2024}
}

@article{eth2020,
author = {Joonho Lee  and Jemin Hwangbo  and Lorenz Wellhausen  and Vladlen Koltun  and Marco Hutter },
title = {Learning quadrupedal locomotion over challenging terrain},
journal = {Science Robotics},
volume = {5},
number = {47},
pages = {eabc5986},
year = {2020},
doi = {10.1126/scirobotics.abc5986},
URL = {https://www.science.org/doi/abs/10.1126/scirobotics.abc5986},
eprint = {https://www.science.org/doi/pdf/10.1126/scirobotics.abc5986},}

@article{parkoureth,
  title={Anymal parkour: Learning agile navigation for quadrupedal robots},
  author={Hoeller, David and Rudin, Nikita and Sako, Dhionis and Hutter, Marco},
  journal={Science Robotics},
  volume={9},
  number={88},
  pages={eadi7566},
  year={2024},
  publisher={American Association for the Advancement of Science}
}

@article{parkourcmu,
title={Extreme Parkour with Legged Robots},
author={Cheng, Xuxin and Shi, Kexin and Agarwal, Ananye and Pathak, Deepak},
journal={arXiv preprint arXiv:2309.14341},
year={2023}
}

@inproceedings{parkourstanford,
  author    = {Zhuang, Ziwen and Fu, Zipeng and Wang, Jianren and Atkeson, Christopher and Schwertfeger, Sören and Finn, Chelsea and Zhao, Hang},
  title     = {Robot Parkour Learning},
  booktitle = {Conference on Robot Learning ({CoRL})},
  year      = {2023},
}

@inproceedings{egocentric,
  title={Legged locomotion in challenging terrains using egocentric vision},
  author={Agarwal, Ananye and Kumar, Ashish and Malik, Jitendra and Pathak, Deepak},
  booktitle={Conference on robot learning},
  pages={403--415},
  year={2023},
  organization={PMLR}
}

@article{asymmetric,
  title={Asymmetric actor critic for image-based robot learning},
  author={Pinto, Lerrel and Andrychowicz, Marcin and Welinder, Peter and Zaremba, Wojciech and Abbeel, Pieter},
  journal={arXiv preprint arXiv:1710.06542},
  year={2017}
}

@article{ppo,
  title={Proximal policy optimization algorithms},
  author={Schulman, John and Wolski, Filip and Dhariwal, Prafulla and Radford, Alec and Klimov, Oleg},
  journal={arXiv preprint arXiv:1707.06347},
  year={2017}
}

@inproceedings{jumpmpc,
  title={Versatile real-time motion synthesis via kino-dynamic mpc with hybrid-systems ddp},
  author={Li, He and Zhang, Tingnan and Yu, Wenhao and Wensing, Patrick M},
  booktitle={2023 IEEE International Conference on Robotics and Automation (ICRA)},
  pages={9988--9994},
  year={2023},
  organization={IEEE}
}

@inproceedings{mopac,
  title={Model predictive actor-critic: Accelerating robot skill acquisition with deep reinforcement learning},
  author={Morgan, Andrew S and Nandha, Daljeet and Chalvatzaki, Georgia and D’Eramo, Carlo and Dollar, Aaron M and Peters, Jan},
  booktitle={2021 IEEE International Conference on Robotics and Automation (ICRA)},
  pages={6672--6678},
  year={2021},
  organization={IEEE}
}

@article{tdmpc2,
  title={Td-mpc2: Scalable, robust world models for continuous control},
  author={Hansen, Nicklas and Su, Hao and Wang, Xiaolong},
  journal={arXiv preprint arXiv:2310.16828},
  year={2023}
}

@inproceedings{dmdmpc,
  title={Dynamic Mirror Descent based Model Predictive Control for Accelerating Robot Learning},
  author={Mishra, Utkarsh A and Samineni, Soumya R and Goel, Prakhar and Kunjeti, Chandravaran and Lodha, Himanshu and Singh, Aman and Sagi, Aditya and Bhatnagar, Shalabh and Kolathaya, Shishir},
  booktitle={2022 International Conference on Robotics and Automation (ICRA)},
  pages={1631--1637},
  year={2022},
  organization={IEEE}
}

@INPROCEEDINGS{rilqr,
  author={Zong, Tongyu and Sun, Liyang and Liu, Yong},
  booktitle={2021 IEEE International Conference on Robotics and Automation (ICRA)}, 
  title={Reinforced iLQR: A Sample-Efficient Robot Locomotion Learning}, 
  year={2021},
  volume={},
  number={},
  pages={5906-5913},
  doi={10.1109/ICRA48506.2021.9561223}}

@inproceedings{d3p,
  title={Making Better Decision by Directly Planning in Continuous Control},
  author={Jinhua Zhu and Yue Wang and Lijun Wu and Tao Qin and Wen-gang Zhou and Tie-Yan Liu and Houqiang Li},
  booktitle={International Conference on Learning Representations},
  year={2023},
  url={https://api.semanticscholar.org/CorpusID:259298829}
}

@misc{isaac_gym,
      title={Isaac Gym: High Performance GPU-Based Physics Simulation For Robot Learning}, 
      author={Viktor Makoviychuk and Lukasz Wawrzyniak and Yunrong Guo and Michelle Lu and Kier Storey and Miles Macklin and David Hoeller and Nikita Rudin and Arthur Allshire and Ankur Handa and Gavriel State},
      year={2021},
      eprint={2108.10470},
      archivePrefix={arXiv},
      primaryClass={cs.RO},
      url={https://arxiv.org/abs/2108.10470}, 
}

@article{pytorch,
  title={Automatic differentiation in PyTorch},
  author={Paszke, Adam and Gross, Sam and Chintala, Soumith and Chanan, Gregory and Yang, Edward and DeVito, Zachary and Lin, Zeming and Desmaison, Alban and Antiga, Luca and Lerer, Adam},
  year={2017}
}

@software{jax,
  author = {James Bradbury and Roy Frostig and Peter Hawkins and Matthew James Johnson and Chris Leary and Dougal Maclaurin and George Necula and Adam Paszke and Jake Vander{P}las and Skye Wanderman-{M}ilne and Qiao Zhang},
  title = {{JAX}: composable transformations of {P}ython+{N}um{P}y programs},
  url = {http://github.com/google/jax},
  version = {0.3.13},
  year = {2018},
}

@misc{dreamer,
      title={Dream to Control: Learning Behaviors by Latent Imagination}, 
      author={Danijar Hafner and Timothy Lillicrap and Jimmy Ba and Mohammad Norouzi},
      year={2020},
      eprint={1912.01603},
      archivePrefix={arXiv},
      primaryClass={cs.LG},
      url={https://arxiv.org/abs/1912.01603}, 
}

@misc{mppi,
      title={Model Predictive Path Integral Control using Covariance Variable Importance Sampling}, 
      author={Grady Williams and Andrew Aldrich and Evangelos Theodorou},
      year={2015},
      eprint={1509.01149},
      archivePrefix={arXiv},
      primaryClass={cs.SY},
      url={https://arxiv.org/abs/1509.01149}, 
}

\end{document}